\documentclass{article}
\usepackage{spconf,amsmath,graphicx}
\usepackage{tikz}
\usepackage{hyperref}

\title{Deep Learning Architectural Designs for \\Super-Resolution of Noisy Images}
\name{Angel Villar-Corrales  \hspace{1cm}  Franziska Schirrmacher  \hspace{1cm}  Christian Riess\thanks{This work was supported in part by the Deutsche Forschungsgemeinschaft (DFG, German Research Foundation) – Project number 146371743 – TRR 89 Invasive Computing and the German Research Foundation, GRK Cybercrime (393541319/GRK2475/1-2019).}}
\address{ IT Security Infrastructures Lab, University of Erlangen-N\"urnberg}
\begin{document}
\ninept
\maketitle
\begin{abstract}
Recent advances in deep learning have led to significant improvements in single image super-resolution (SR) research.
However, due to the amplification of noise during the upsampling steps, state-of-the-art methods often fail at reconstructing high-resolution images from noisy versions of their low-resolution counterparts. However, this is especially important for images from unknown cameras with unseen types of image degradation.
In this work, we propose to jointly perform denoising and super-resolution. To this end, we investigate two architectural designs: ``in-network'' combines both tasks at feature level, while ``pre-network'' first performs denoising and then super-resolution.
Our experiments show that both variants have specific advantages:
The in-network design obtains the strongest results when the type of image corruption is aligned in the training and testing dataset, for any choice of denoiser.
The pre-network design exhibits superior performance on unseen types of image corruption, which is a pathological failure case of existing super-resolution models. We hope that these findings help to enable super-resolution also in less constrained scenarios where source camera or imaging conditions are not well controlled.
Source code and pretrained models are available at \url{https://github.com/angelvillar96/super-resolution-noisy-images}.
\end{abstract}

\section{Introduction}

Single image super-resolution (SR) aims at recovering a high-resolution (HR) image from its low-resolution (LR) counterpart, in which high-frequency details have been lost due to degrading factors such as blur, hardware limitations, or decimation. 

Early SR approaches were based on upsampling and interpolation techniques~\cite{li2001new,zhang2006edge}.
However, these methods are limited in their representational power, and hence also limited in their ability to predict realistic high-resolution images. More complex methods construct mapping functions between low- and high-resolution images.
Such a mapping function can be obtained from a variety of different techniques, including sparse coding~\cite{zeyde2010single,yang2010image}, random forests~\cite{salvador2015naive,schulter2015fast} or embedding approaches~\cite{timofte2013anchored,timofte2014a+}. Recently, deep learning methods for super-resolution lead to considerable performance improvements \cite{dong2014learning, kim2016deeply}. ResNet-like architectures~\cite{he2016deep} obtain state-of-the-art results for SR tasks while maintaining low computational complexity~\cite{Lim_EnhancedResidualSR_2017,Yu_WideActivationSuperResolution_2018, Zhang_ResidualSR_2018}.
Despite these successes, it is still challenging to prevent the amplification of noise during the upsampling steps, which often leads to loss of information and the emergence of artifacts.

\begin{figure}[t]
	\centering
	\includegraphics[width=\linewidth]{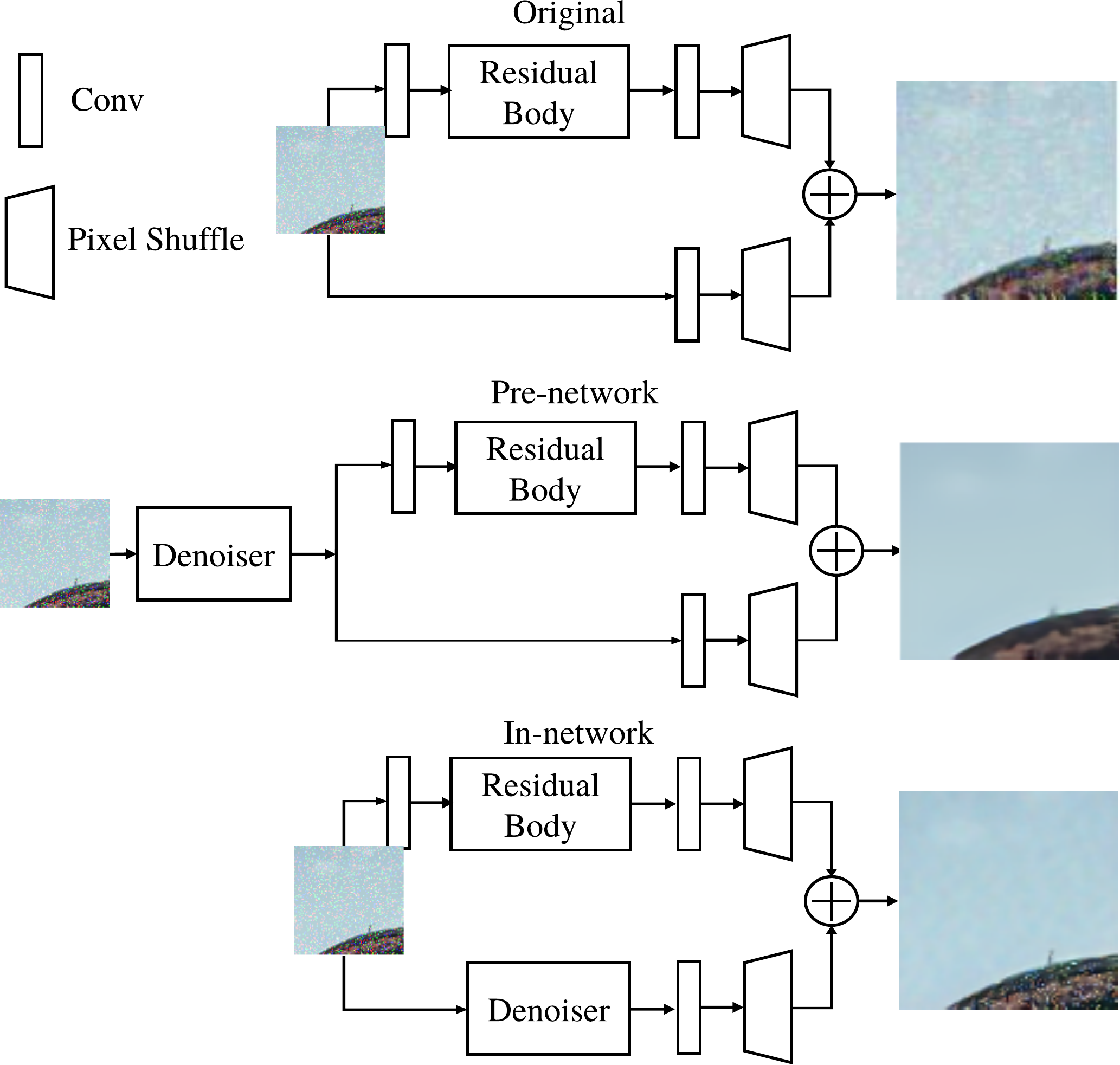}
	\caption{Original WDSR architecture (top) in comparison to the pre-network architectural design (center) and the in-network architectural design (bottom). Pre-network cascades denoising and super-resolution. In-network combines low-level features from the denoised input with high-level features from the noisy input.}
	\label{fig:model architecture}
\end{figure}

Several approaches have been considered to jointly perform super-resolution and denoising. 
Image restoration can be formulated as an inverse problem. In this approach, the data term for the respective objectives is specific to the respective task. For the prior, a more generic function can be chosen that applies to multiple tasks. For example, using deep learning models~\cite{Tirer2019}, employing a denoiser as regularizer~\cite{romano2017little,schirrmacher2020adaptive} or a so-called plug-and-play prior~\cite{chan2016plug}. Furthermore, deep learning approaches have also been considered to combine denoising with a SR model \cite{zhang2017_CnnDenoiserPrior}. The authors in~\cite{Bei_DenoisingSuperResolution_2018} propose cascading a denoiser with a SR model so that the output of the denoiser is fed to the SR network.  The pre-network architectural design in our experiments is similar in spirit, but instead of using a fixed convolutional neural network (CNN) denoiser, our design allows for further flexibility regarding the choice of the integrated denoiser. 

In contrast to previous works, we compare two architectures that allow for further flexibility regarding the choice of the integrated denoiser.
This flexibility can be used to incorporate domain knowledge into the network by selecting a denoising technique optimized for the particular type of degradation. For scenarios where domain knowledge is missing, we investigate the generalization capability of different denoisers and the proposed architectural designs. Especially for low-resolution images from cameras in-the-wild, image degradation, such as unseen noise distributions, can lead to artifacts in the reconstructed high-resolution images.

The first architecture, ``pre-network'', includes a denoiser as a preprocessing step to the super-resolution. The second architecture, ``in-network'', reconstructs the HR image by combining low-level features extracted from the denoised input and high-level features extracted from the noisy input.  To the best of our knowledge, this type of network design has not yet been studied in the context of denoising and super-resolution.

We evaluate our architectures with the Wide Activation Super-Resolution model (WDSR)~\cite{Yu_WideActivationSuperResolution_2018} on noisy versions of the images from the DIV2K~\cite{Agustsson_NRIREChallenge_2017} dataset. Both architectures aim to suppress the noise corruptions while still recovering most of the high-frequency information. We show that both architectures achieve more realistic reconstructions and better PSNR values than WDSR only.

	
	

\section{Methods}\label{sec:methods}


In this work, we employ the Wide Activation Super-Resolution (WDSR) model~\cite{Yu_WideActivationSuperResolution_2018}  as a building block to investigate architectures for joint denoising and super-resolution.

The original WDSR architecture is shown on top of Figure~\ref{fig:model architecture}. It consists of two paths. The main path is on top, consisting of a user-defined number $B$ of residual blocks. Each block consists of two convolutional layers followed by weight normalization~\cite{Salimans_WeightNormalization_2016} and ReLU activation. The lower path is a residual connection. It provides low-level features from the input to the output, which is critically important for SR tasks~\cite{Zhang_ResidualSR_2018}. Both paths contain a pixel-shuffle layer~\cite{Shi_PixelShuffleSR_2016} 
, which performs the upsampling for image super-resolution.
WDSR is sensitive to input images that are corrupted by additive noise. However, we show in this work that it can be paired with a denoiser in two different ways, denoted as ``pre-network'' and ``in-network'', which both considerably improve the results.

\begin{table*}[t!]
	\renewcommand{\arraystretch}{1}
	\caption{Average peak signal-to-noise ratio (PSNR) for varying levels of additive Gaussian noise $\mathbf{\sigma^2}$ on the low-resolution testing images. Baseline WDSR fine-tuned on the noisy data (``No denoiser'') performs best at low noise powers, while the in-net with an autoencoder denoiser achieves best the PSNR values for $\sigma^2 \geq 0.20$. Baseline WDSR without fine-tuning (``No tuning'') performs worst (see text for details).}
	\label{table:evaluations div2k gaussian}
	\centering
	\begin{tabular}{lrlrlrllrllrll}
		\hline
		&&  && && \multicolumn{2}{c}{\textbf{Median Filter}} && \multicolumn{2}{c}{\textbf{Wiener Filter}} && \multicolumn{2}{c}{\textbf{Autoencoder}}\\
		$\mathbf{\sigma^2}$ && \textbf{No tuning} && \textbf{No denoiser} && \textbf{In-net} & \textbf{Pre-net} && \textbf{In-net} & \textbf{Pre-net}  && \textbf{In-net} & \textbf{Pre-net}\\ 
		\hline\hline
		
		0.00 && 32.16 &&  32.29 && \textbf{32.62}  &  29.23  &&  32.37 &  28.95  &&  32.15 &  24.58  \\
		
		0.05 && 23.54 &&  \textbf{29.99} && 29.76  &  28.30  &&  29.32 &  27.26  &&  29.74 &  24.02 \\ 
		
		0.10 && 18.78 &&  \textbf{28.28} && 28.09  &  27.02  &&  27.73 &  26.31  &&  28.04 &  23.94  \\
		
		0.15 && 15.79 &&  \textbf{27.00} && 26.67  &  26.03  &&  26.34 &  25.69  &&  26.69 &  23.24  \\
		
		0.20 && 13.81 &&  25.28 && 25.54  &  25.32  &&  25.45 &  25.08  &&  \textbf{25.88} &  23.05  \\
		
		0.25 && 12.40 &&  25.12 && 24.90  &  24.83  &&  24.83 &  24.33  &&  \textbf{25.64} &  22.85  \\
		
		0.30 && 11.32 &&  24.45 && 24.40  &  24.26  &&  24.34 &  23.67  &&  \textbf{24.64} &  22.86  \\
		
		\hline
	\end{tabular}
\end{table*}

Figure~\ref{fig:model architecture} shows both architectures.
Pre-network (abbreviated pre-net) is shown in the middle. It first passes the image through the denoiser 
prior to branching into main path and skip connection. This is conceptually similar to the denoiser and SR concatenation by 
Bei~\emph{et al.}~\cite{Bei_DenoisingSuperResolution_2018}.
One potential limitation of this approach is error propagation:
if the denoiser removes information that is relevant to super-resolution, it cannot be recovered afterwards.

In-network (abbreviated in-net) is shown on the bottom of Fig.~\ref{fig:model architecture}.
Here, the denoiser integrates into the residual connection. Hence, the SR model can jointly combine
low-level features from the denoised input and high-level features from the noisy input.

Both designs are open for the choice of denoiser, which allows to choose a task-specific denoiser, i.e.,
that performs best on an expected noise distribution.
In our experiments, we evaluate three popular denoisers of varying complexity: median filter~\cite{Fan_ReviewDenoising_2019}, Wiener filter~\cite{Wiener_WienerFilter_1964}, and denoising autoencoders (DAE)~\cite{Vincent_DenoisingAutoencoders_2010}.

\begin{figure*}[t]
    \centering
  		\begin{tikzpicture}	  
        	    \node at (0,0.1) {\includegraphics[width=.38\textwidth]{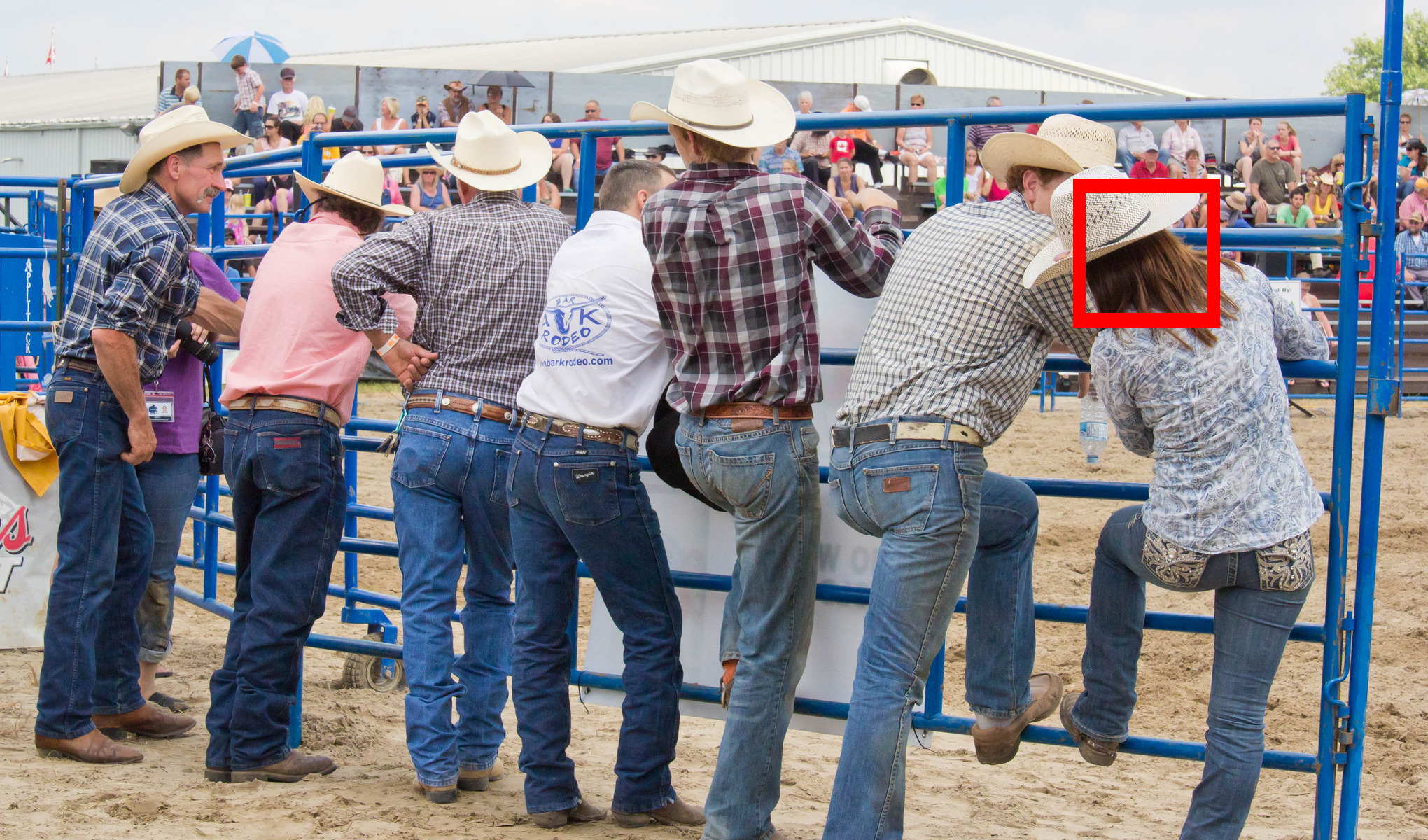}};
        	    \node at (4.5,1.35) {\includegraphics[width=.12\textwidth]{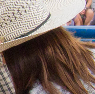}};
                \node at (4.5,2.55)[] {Original};
                \node at (4.5,-1.15)
                 {\includegraphics[width=.12\textwidth]{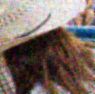}};
                \node at (4.5,0.05)[] {No tuning};
                
        	    \node at (6.7,1.35) {\includegraphics[width=.12\textwidth]{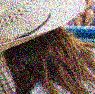}};
                \node at (6.7,2.55)[] {Input};
                \node at (6.7,-1.15)
                 {\includegraphics[width=.12\textwidth]{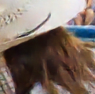}};
                \node at (6.7,0.05)[] {No denoiser};
                
                \node at (8.9,1.35) {\includegraphics[width=.12\textwidth]{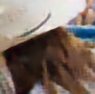}};
                \node at (8.9,2.55)[] {Median pre-net};
                \node at (8.9,-1.15)
                 {\includegraphics[width=.12\textwidth]{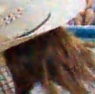}};
                \node at (8.9,0.05)[] {Median in-net};
                
                \node at (11.1,1.35) {\includegraphics[width=.12\textwidth]{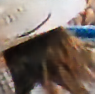}};
                \node at (11.1,2.55)[] {Wiener pre-net};
                \node at (11.1,-1.15)
                 {\includegraphics[width=.12\textwidth]{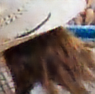}};
                \node at (11.1,0.05)[] {Wiener in-net};
                
                \node at (13.3,1.35) {\includegraphics[width=.12\textwidth]{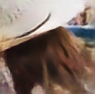}};
                \node at (13.3,2.55)[] {DAE pre-net};
                \node at (13.3,-1.15)
                 {\includegraphics[width=.12\textwidth]{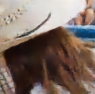}};
                \node at (13.3,0.05)[] {DAE in-net};
    	    \end{tikzpicture}
    \caption{Comparison of different reconstructions of an image patch corrupted with Gaussian noise with $\sigma^2=0.15$ using different denoisers and architectural designs. ``No tuning'' corresponds to the reconstruction using the original WDSR trained  on clean images and ``No denoiser'' corresponds to the reconstruction using a WDSR model fine-tuned on noisy patches.}
    \label{fig:image comparisons}
\end{figure*}
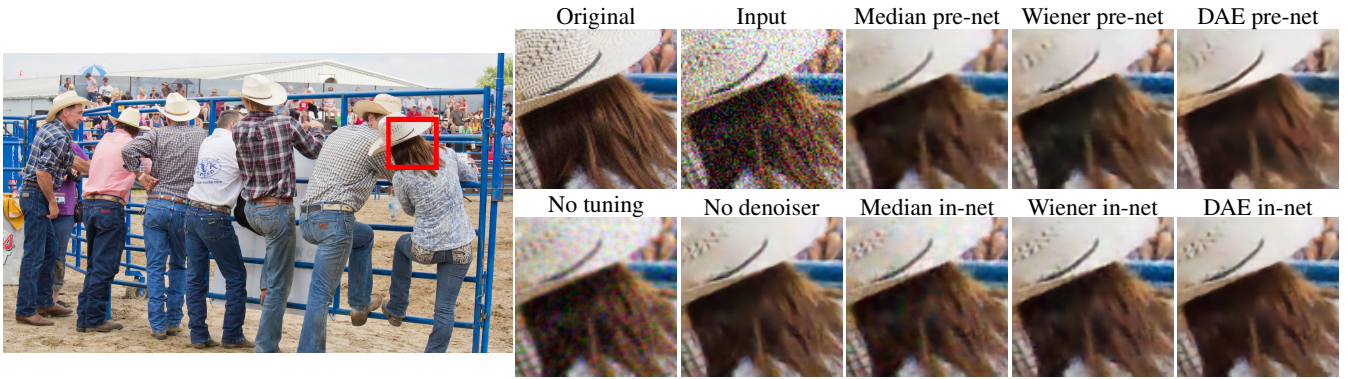


\section{Dataset Preparation and Training Procedure}\label{sec:training}

The experiments use the DIV2K dataset. In our experiments, we consider low-resolution images which have been downsampled by a factor of two. 
We use the 800 images from the training set for training or fine-tuning the models, and the 100 validation images for evaluation. Closely following \cite{Yu_WideActivationSuperResolution_2018}, we feed to the model $96 \times 96$ RGB image patches extracted from HR images, along with their noisy bicubic downsampled counterparts.

We train our models using downsampled image patches corrupted with additive Gaussian noise. The testing data is corrupted using additive Gaussian noise with the same distribution as during training. 
Moreover, Poisson, speckle, and salt-and-pepper noise are used to degrade the testing data for the robustness analysis.

As a baseline, we consider two versions of WDSR. First, the publicly available version of WDSR, pretrained on the DIV2K dataset, denoted as ``No tuning''. Second, WDSR fine-tuned on DIV2K images with added noise, denoted as ``No denoiser''.
The model weights are initialized with the pretrained WDSR weights. Each model is fine-tuned using the mean absolute error (MAE) loss function and the ADAM update rule~\cite{Kingma_Adam_2014} with an initial learning rate of $10^{-4}$. To avoid overfitting, the fine-tuning procedure is stopped after 100 epochs.
Regarding the median and Wiener filter, we use square kernel sizes with side length of five pixels.
For the denoising autoencoders, we construct a fully convolutional DAE composed of three convolutional layers with 64, 128, and 256 kernels of size $5 \times 5$ respectively. Each layer is followed by a ReLU activation function and max-pooling. DAEs are trained for 80 epochs using the MSE loss function and the ADAM update rule with an initial learning rate of $10^{-4}$. For the integration into the WDSR model, the autoencoder parameters are fixed during the fine-tuning of the network.


\begin{figure*}[t]
    \centering
  		\begin{tikzpicture}	  
        	    \node at (0,0.1) {\includegraphics[width=.38\textwidth]{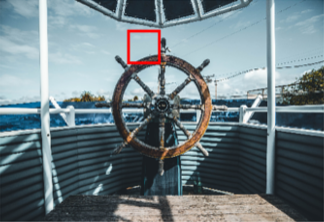}};
        	    \node at (4.5,1.35) {\includegraphics[width=.12\textwidth]{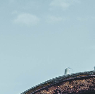}};
                \node at (4.5,2.55)[] {Original};
                \node at (4.5,-1.15)
                 {\includegraphics[width=.12\textwidth]{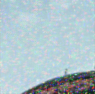}};
                \node at (4.5,0.05)[] {No tuning};
                
        	    \node at (6.7,1.35) {\includegraphics[width=.12\textwidth]{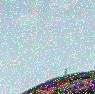}};
                \node at (6.7,2.55)[] {Input};
                \node at (6.7,-1.15)
                 {\includegraphics[width=.12\textwidth]{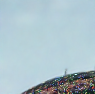}};
                \node at (6.7,0.05)[] {No denoiser};
                
                \node at (8.9,1.35) {\includegraphics[width=.12\textwidth]{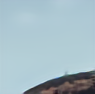}};
                \node at (8.9,2.55)[] {Median pre-net};
                \node at (8.9,-1.15)
                 {\includegraphics[width=.12\textwidth]{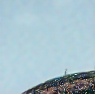}};
                \node at (8.9,0.05)[] {Median in-net};
                
                \node at (11.1,1.35) {\includegraphics[width=.12\textwidth]{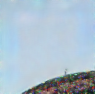}};
                \node at (11.1,2.55)[] {Wiener pre-net};
                \node at (11.1,-1.15)
                 {\includegraphics[width=.12\textwidth]{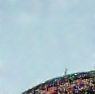}};
                \node at (11.1,0.05)[] {Wiener in-net};
                
                \node at (13.3,1.35) {\includegraphics[width=.12\textwidth]{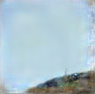}};
                \node at (13.3,2.55)[] {DAE pre-net};
                \node at (13.3,-1.15)
                 {\includegraphics[width=.12\textwidth]{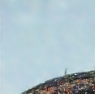}};
                \node at (13.3,0.05)[] {DAE in-net};
    	    \end{tikzpicture}
    \caption{Comparison of different reconstructions of an image patch with Poisson noise using different denoisers and architectures. ``No tuning'' corresponds to the reconstruction using the original WDSR trained on clean images and ``No denoiser'' corresponds to the reconstruction using a WDSR model fine-tuned on noisy patches.}
    \label{fig:image comparisons poisson}
\end{figure*}


\begin{table*}[t]
	\renewcommand{\arraystretch}{1.}
	\caption{Average PSNR values for noise distributions that differ from the training distribution. As a baseline, Gaussian noise is aligned with training distribution. Here, and for the mathematically similar speckle noise, WDSR without a denoiser (``No denoiser'') performs best. For the more challenging salt and pepper (S\&P) noise and Poisson noise, the pre-network again generalizes best, by a large margin.}
	\label{table:evaluations generalization noise}
	\centering
	\begin{tabular}{llrlrllrllrll}
		\hline
		&  && && \multicolumn{2}{c}{\textbf{Median Filter}} && \multicolumn{2}{c}{\textbf{Wiener Filter}} && \multicolumn{2}{c}{\textbf{Autoencoder}}\\
		Noise & \textbf{No tuning} && \textbf{No denoiser} && \textbf{In-net} & \textbf{Pre-net} && \textbf{In-net} & \textbf{Pre-net}  && \textbf{In-net} & \textbf{Pre-net}\\ 
		\hline\hline
		
		Gaussian & 18.78 && \textbf{28.28} && 28.09  &  27.02  &&  27.73 &  26.31  &&  28.04 &  23.94  \\
		
		Speckle  & 24.28 && \textbf{27.86} && 27.63  &  26.78  &&  26.88 &  24.48  &&  27.54 &  23.20  \\
		
		Poisson  & 14.34 && 19.47 && 21.30  &  \textbf{26.69}  &&  20.49 &  20.50  &&  20.59 &  18.16  \\ 
		
		S\&P     & 10.78 && 12.77 && 13.45  &  \textbf{26.32}  &&  14.79 &  16.57  &&  12.76 &  13.82  \\
		\hline
	\end{tabular}
\end{table*}

\section{Evaluation}\label{sec:evaluation}

\begin{figure*}[t!]
    \centering
  		\begin{tikzpicture}	  
        	    \node at (0,0.1) {\includegraphics[width=.35\textwidth]{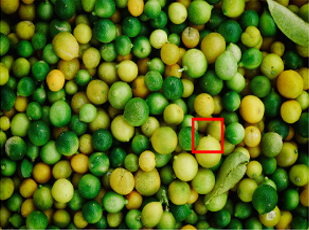}};
        	    \node at (4.5,1.35) {\includegraphics[width=.12\textwidth]{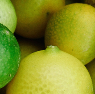}};
                \node at (4.5,2.55)[] {Original};
                \node at (4.5,-1.15)
                 {\includegraphics[width=.12\textwidth]{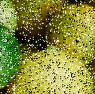}};
                \node at (4.5,0.05)[] {No tuning};
                
        	    \node at (6.7,1.35) {\includegraphics[width=.12\textwidth]{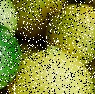}};
                \node at (6.7,2.55)[] {Input};
                \node at (6.7,-1.15)
                 {\includegraphics[width=.12\textwidth]{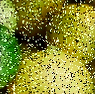}};
                \node at (6.7,0.05)[] {No denoiser};
                
                \node at (8.9,1.35) {\includegraphics[width=.12\textwidth]{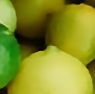}};
                \node at (8.9,2.55)[] {Median pre-net};
                \node at (8.9,-1.15)
                 {\includegraphics[width=.12\textwidth]{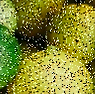}};
                \node at (8.9,0.05)[] {Median in-net};
                
                \node at (11.1,1.35) {\includegraphics[width=.12\textwidth]{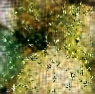}};
                \node at (11.1,2.55)[] {Wiener pre-net};
                \node at (11.1,-1.15)
                 {\includegraphics[width=.12\textwidth]{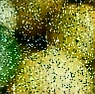}};
                \node at (11.1,0.05)[] {Wiener in-net};
                
                \node at (13.3,1.35) {\includegraphics[width=.12\textwidth]{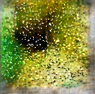}};
                \node at (13.3,2.55)[] {DAE pre-net};
                \node at (13.3,-1.15)
                 {\includegraphics[width=.12\textwidth]{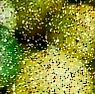}};
                \node at (13.3,0.05)[] {DAE in-net};
    	    \end{tikzpicture}
    \caption{Comparison of reconstructions of an image patch with salt-and-pepper noise using different denoisers and architectures. ``No tuning'' corresponds to the reconstruction using the original WDSR trained on clean images and ``No denoiser'' corresponds to the reconstruction using WDSR fine-tuned on noisy patches. Only the pre-network with the median filter succeeds at denoising and reconstructing the HR image.}
    \label{fig:image comparisons salt-and-pepper}
\end{figure*}


\subsection{Aligned Noise in Training and Testing}\label{sec:eval_aligned}

The noise in real images
can often be approximated as additive Gaussian noise,
which we model in this experiment as a fixed power $\sigma^2$ for training and testing.
This evaluation assumes that noise distribution and strength are known during training. This is a strong assumption for practical cases, but this setup shows the fundamental capability of the models to capture non-ideal images.


Table~\ref{table:evaluations div2k gaussian} shows the numerical performances for the peak signal-to-noise ratio (PSNR). The experiment is performed for additive Gaussian noise with $0.00~\le~\sigma^2~\le~0.30$. For low noise levels, the fine-tuned WDSR (``No denoiser'') and in-network with Median filter perform best. In-network with the denoising autoencoder outperforms the competing methods for higher noise levels.
A consistent decrease in PSNR can be observed with increasing strength of the noise. This is expected, since image restoration becomes increasingly difficult with increasing noise strength.
Pre-net and in-net are both affected by increasing noise, since the denoiser increasingly removes useful information from the image.

In comparison, the original WDSR model (``No tuning'') is unable to accurately recover the HR image from its noisy LR counterpart. However, fine-tuning WDSR (``No denoiser'') considerably improves the PSNR, for example by about 10~dB for $\sigma^2 = 0.1$.
Thus, the fine-tuned WDSR without any integrated denoiser learns to suppress noise and is on par with the best denoiser methods (i.e., in-net with autoencoder and median filter).
Hence, WDSR's 16 residual blocks and 32 convolutional filters apparently possess
sufficient representational power
to jointly learn denoising and SR. However, we will show in the next section that these findings do not generalize to distortions that were not seen during training.

Figure~\ref{fig:image comparisons} shows a qualitative comparison of the reconstructions of an image patch for a noise strength of $\sigma^2=0.1$. The in-network architecture yields very similar results for all denoisers. It exhibits comparable results to the fine-tuned WDSR and overall outperforms pre-network. Upon closer examination, pre-net suffers from the denoiser's removal of relevant information at a very early processing state. This can be seen in the holes in the hat. The holes appear blurry or are completely removed in all pre-net configurations. In-net performs much better since it complements the denoised image in the skip branch with original image information in the main branch.

In conclusion, when training and testing images are aligned in a fixed amount of Gaussian noise, the fine-tuned WDSR without any denoiser and the in-net exhibit the best PSNR results and generate the most accurate reconstructions.

\subsection{Generalization to Unseen Noise Distributions}\label{sec:eval_distributions}

In this experiment, we investigate the generalization to noise distributions that were not part of the training data.
The models are trained on images with additive Gaussian noise of strength $\sigma^2=0.1$. Testing is performed on three different noise distributions.
More specifically, we use speckle noise with $\sigma^2=0.1$,  Poisson noise with $\lambda=0.1$, and salt-and-pepper (S\&P) noise with a probability of $p=0.2$ to change a pixel.

Table~\ref{table:evaluations generalization noise} shows quantitative results for this experiment.
As a reference, we also include performances for testing on additive Gaussian noise with $\sigma^2 = 0.1$, i.e., where training and testing are aligned. The fine-tuned WDSR (``No denoiser'') performs best on images corrupted by Gaussian noise and speckle noise.
We attribute this to the fact that speckle noise is a multiplicative distortion, but it follows the same probabilistic distribution as additive Gaussian noise. 
In-network with the median filter and the denoising autoencoder achieves similar results. However, pre-network with the median filter outperforms the competing methods by a large margin on images that are corrupted by salt-and-pepper noise and Poisson noise: an improvement over the baselines of more than 13~dB is achieved. 

For noise distributions that significantly differ from those seen during training, it is highly beneficial to suppress as much of the input corruption as possible prior to performing SR. For this reason, pre-network performs best on Poisson noise and salt-and-pepper noise.
The choice of the denoiser also affects the results.
To this end, Fig.~\ref{fig:image comparisons poisson} shows qualitative results on Poisson noise.
All configurations of in-net, as well as pre-net with Wiener filter, and the fine-tuned WDSR (``No denoiser'') suppress noise in homogeneous areas, such as the blue sky.
However, the wheel exhibits more texture, and it still contains noise.
Pre-net with median filter removes the noise best, but the reconstructed image appears slightly over-smoothed.

In Fig.~\ref{fig:image comparisons salt-and-pepper}, qualitative results on salt-and-pepper noise are shown.
Here,  the median filter achieves the best reconstructions. The noise is completely removed, but some details of the lemon peel are also removed. We attribute both observations to to the fact that median filtering is particularly well-suited for suppressing salt-and-pepper noise but does not preserve small structures. The Wiener filter in the pre-net configuration generates unwanted artifacts since it is optimal for Gaussian denoising. Finally, it is interesting to note that the PSNR of the denoising autoencoder is oftentimes outperformed by the other denoisers.
This can be attributed to the fact that the autoencoder is trained on additive Gaussian noise, and it apparently has difficulties to generalize to other distributions without retraining. 

In summary, models that are trained on additive Gaussian noise do not generalize well beyond Gaussian and speckle noise. Pre-network combined with a suitable denoiser generalizes considerably better to different noise distributions than the competing methods.

\section{Conclusion}\label{sec:conclusion}

In this paper, we compare two architectures to jointly perform image denoising and single-image super-resolution. We combine the well-known WDSR model~\cite{Yu_WideActivationSuperResolution_2018} with three denoisers that can be chosen depending on the type of degradation. Both networks have specific benefits. The ``pre-network'' architecture sequentially removes the noise first, and then recovers the high-resolution image. With a suitable denoiser, pre-network generalizes well to unseen noise distributions. However, details in the image are removed and the reconstructions appear slightly over-smoothed.
The ``in-network'' architecture reconstructs the high-resolution image by combining low-level features from the denoiser with high-level features from the noisy input. This enables better structure preservation and sharper reconstructed images, but is more sensitive to unseen noise distributions and strength, independent of the chosen denoiser. We hope that these findings are useful toward enabling super-resolution in-the-wild, when camera and image conditions are not fully controlled.


%
%



\bibliographystyle{IEEEbib}
\bibliography{references}

\begin{thebibliography}{10}

\bibitem{li2001new}
Xin Li and Michael~T Orchard,
\newblock ``New edge-directed interpolation,''
\newblock {\em IEEE transactions on image processing}, vol. 10, no. 10, pp.
  1521--1527, 2001.

\bibitem{zhang2006edge}
Lei Zhang and Xiaolin Wu,
\newblock ``An edge-guided image interpolation algorithm via directional
  filtering and data fusion,''
\newblock {\em IEEE transactions on Image Processing}, vol. 15, no. 8, pp.
  2226--2238, 2006.

\bibitem{zeyde2010single}
Roman Zeyde, Michael Elad, and Matan Protter,
\newblock ``On single image scale-up using sparse-representations,''
\newblock in {\em International conference on curves and surfaces}. Springer,
  2010, pp. 711--730.

\bibitem{yang2010image}
Jianchao Yang, John Wright, Thomas~S Huang, and Yi~Ma,
\newblock ``Image super-resolution via sparse representation,''
\newblock {\em IEEE transactions on image processing}, vol. 19, no. 11, pp.
  2861--2873, 2010.

\bibitem{salvador2015naive}
Jordi Salvador and Eduardo Perez-Pellitero,
\newblock ``Naive bayes super-resolution forest,''
\newblock in {\em Proceedings of the IEEE International conference on computer
  vision}, 2015, pp. 325--333.

\bibitem{schulter2015fast}
Samuel Schulter, Christian Leistner, and Horst Bischof,
\newblock ``Fast and accurate image upscaling with super-resolution forests,''
\newblock in {\em Proceedings of the IEEE Conference on Computer Vision and
  Pattern Recognition}, 2015, pp. 3791--3799.

\bibitem{timofte2013anchored}
Radu Timofte, Vincent De~Smet, and Luc Van~Gool,
\newblock ``Anchored neighborhood regression for fast example-based
  super-resolution,''
\newblock in {\em Proceedings of the IEEE international conference on computer
  vision}, 2013, pp. 1920--1927.

\bibitem{timofte2014a+}
Radu Timofte, Vincent De~Smet, and Luc Van~Gool,
\newblock ``A+: Adjusted anchored neighborhood regression for fast
  super-resolution,''
\newblock in {\em Asian conference on computer vision}. Springer, 2014, pp.
  111--126.

\bibitem{dong2014learning}
Chao Dong, Chen~Change Loy, Kaiming He, and Xiaoou Tang,
\newblock ``Learning a deep convolutional network for image super-resolution,''
\newblock in {\em European conference on computer vision}. Springer, 2014, pp.
  184--199.

\bibitem{kim2016deeply}
Jiwon Kim, Jung Kwon~Lee, and Kyoung Mu~Lee,
\newblock ``Deeply-recursive convolutional network for image
  super-resolution,''
\newblock in {\em Proceedings of the IEEE conference on computer vision and
  pattern recognition}, 2016, pp. 1637--1645.

\bibitem{he2016deep}
Kaiming He, Xiangyu Zhang, Shaoqing Ren, and Jian Sun,
\newblock ``Deep residual learning for image recognition,''
\newblock in {\em Proceedings of the IEEE conference on computer vision and
  pattern recognition}, 2016, pp. 770--778.

\bibitem{Lim_EnhancedResidualSR_2017}
Bee Lim, Sanghyun Son, Heewon Kim, Seungjun Nah, and Kyoung Mu~Lee,
\newblock ``Enhanced deep residual networks for single image
  super-resolution,''
\newblock in {\em Proceedings of the IEEE conference on computer vision and
  pattern recognition workshops}, 2017, pp. 136--144.

\bibitem{Yu_WideActivationSuperResolution_2018}
Jiahui Yu, Yuchen Fan, Jianchao Yang, Ning Xu, Zhaowen Wang, Xinchao Wang, and
  Thomas Huang,
\newblock ``Wide activation for efficient and accurate image
  super-resolution,''
\newblock {\em arXiv preprint arXiv:1808.08718}, 2018.

\bibitem{Zhang_ResidualSR_2018}
Yulun Zhang, Yapeng Tian, Yu~Kong, Bineng Zhong, and Yun Fu,
\newblock ``Residual dense network for image super-resolution,''
\newblock in {\em Proceedings of the IEEE conference on computer vision and
  pattern recognition}, 2018, pp. 2472--2481.

\bibitem{Tirer2019}
Tom Tirer and Raja Giryes,
\newblock ``Super-resolution via image-adapted denoising cnns: Incorporating
  external and internal learning,''
\newblock {\em IEEE Signal Processing Letters}, vol. 26, no. 7, pp. 1080--1084,
  2019.

\bibitem{romano2017little}
Yaniv Romano, Michael Elad, and Peyman Milanfar,
\newblock ``The little engine that could: Regularization by denoising (red),''
\newblock {\em SIAM Journal on Imaging Sciences}, vol. 10, no. 4, pp.
  1804--1844, 2017.

\bibitem{schirrmacher2020adaptive}
Franziska Schirrmacher, Christian Riess, and Thomas K{\"o}hler,
\newblock ``Adaptive quantile sparse image (aquasi) prior for inverse imaging
  problems,''
\newblock {\em IEEE Transactions on Computational Imaging}, vol. 6, pp.
  503--517, 2020.

\bibitem{chan2016plug}
Stanley~H Chan, Xiran Wang, and Omar~A Elgendy,
\newblock ``Plug-and-play admm for image restoration: Fixed-point convergence
  and applications,''
\newblock {\em IEEE Transactions on Computational Imaging}, vol. 3, no. 1, pp.
  84--98, 2016.

\bibitem{zhang2017_CnnDenoiserPrior}
Kai Zhang, Wangmeng Zuo, Shuhang Gu, and Lei Zhang,
\newblock ``Learning deep cnn denoiser prior for image restoration,''
\newblock in {\em Proceedings of the IEEE conference on computer vision and
  pattern recognition}, 2017, pp. 3929--3938.

\bibitem{Bei_DenoisingSuperResolution_2018}
Yijie Bei, Alexandru Damian, Shijia Hu, Sachit Menon, Nikhil Ravi, and Cynthia
  Rudin,
\newblock ``New techniques for preserving global structure and denoising with
  low information loss in single-image super-resolution,''
\newblock in {\em Proceedings of the IEEE Conference on Computer Vision and
  Pattern Recognition Workshops}, 2018, pp. 874--881.

\bibitem{Agustsson_NRIREChallenge_2017}
Radu Timofte, Eirikur Agustsson, Luc Van~Gool, Ming-Hsuan Yang, and Lei Zhang,
\newblock ``Ntire 2017 challenge on single image super-resolution: Methods and
  results,''
\newblock in {\em Proceedings of the IEEE conference on computer vision and
  pattern recognition workshops}, 2017, pp. 114--125.

\bibitem{Salimans_WeightNormalization_2016}
Tim Salimans and Durk~P Kingma,
\newblock ``Weight normalization: A simple reparameterization to accelerate
  training of deep neural networks,''
\newblock in {\em Advances in neural information processing systems}, 2016, pp.
  901--909.

\bibitem{Shi_PixelShuffleSR_2016}
Wenzhe Shi, Jose Caballero, Ferenc Husz{\'a}r, Johannes Totz, Andrew~P Aitken,
  Rob Bishop, Daniel Rueckert, and Zehan Wang,
\newblock ``Real-time single image and video super-resolution using an
  efficient sub-pixel convolutional neural network,''
\newblock in {\em Proceedings of the IEEE conference on computer vision and
  pattern recognition}, 2016, pp. 1874--1883.

\bibitem{Fan_ReviewDenoising_2019}
Linwei Fan, Fan Zhang, Hui Fan, and Caiming Zhang,
\newblock ``Brief review of image denoising techniques,''
\newblock {\em Visual Computing for Industry, Biomedicine, and Art}, vol. 2,
  no. 1, pp. 7, 2019.

\bibitem{Wiener_WienerFilter_1964}
Norbert Wiener,
\newblock {\em Extrapolation, interpolation, and smoothing of stationary time
  series},
\newblock The MIT press, 1964.

\bibitem{Vincent_DenoisingAutoencoders_2010}
Pascal Vincent, Hugo Larochelle, Isabelle Lajoie, Yoshua Bengio, and
  Pierre-Antoine Manzagol,
\newblock ``Stacked denoising autoencoders: Learning useful representations in
  a deep network with a local denoising criterion,''
\newblock {\em Journal of machine learning research}, vol. 11, no. Dec, pp.
  3371--3408, 2010.

\bibitem{Kingma_Adam_2014}
Diederik~P Kingma and Jimmy Ba,
\newblock ``Adam: A method for stochastic optimization,''
\newblock {\em arXiv preprint arXiv:1412.6980}, 2014.

\end{thebibliography}

\end{document}